\documentclass[10pt,twocolumn,letterpaper]{article}

\usepackage{cvpr}      

\usepackage{algorithm}
\usepackage{algorithmic}

\usepackage[utf8]{inputenc} 
\usepackage[T1]{fontenc}    
\usepackage{amsmath}
\usepackage{amsthm}

\usepackage{url}            
\usepackage{booktabs}       
\usepackage{amsfonts}       
\usepackage{nicefrac}       
\usepackage{microtype}      
\usepackage{xcolor}         
\usepackage[table]{xcolor}
\usepackage{graphicx}
\usepackage{diagbox}
\usepackage{url}
\usepackage{booktabs}
\usepackage{bbding}
\usepackage{pifont}
\usepackage{wasysym}
\usepackage{utfsym}
\usepackage{fontawesome}
\usepackage{appendix}
\usepackage{multirow}
\usepackage{enumitem}
\usepackage[breakable]{tcolorbox}
\usepackage{tabularx}
\usepackage{makecell}

\definecolor{cvprblue}{rgb}{0.21,0.49,0.74}
\usepackage[pagebackref,breaklinks,colorlinks,allcolors=cvprblue]{hyperref}

\def\confName{CVPR}
\def\confYear{2027}

\title{Breaking Visual Conventions: \\Spectrally Coupled Weight Perturbation for Creative Image Generation}

\author{Fu Feng\textsuperscript{\rm 1,2}\thanks{Equal contribution}\;\; Ruoheng Zhou\textsuperscript{\rm 1,2}\footnotemark[1]\;\; Yucheng Xie\textsuperscript{\rm 1,2}\;\; Xu Yang\textsuperscript{\rm 1,2}\;\; Jing Wang\textsuperscript{\rm 1,2}\thanks{Corresponding authors}\;\;\; Xin Geng\textsuperscript{\rm 1,2}\footnotemark[2]\\
\textsuperscript{\rm 1}School of Computer Science and Engineering, Southeast University, Nanjing, China\\
\textsuperscript{\rm 2}Key Laboratory of New Generation Artificial Intelligence Technology and Its Interdisciplinary \\Applications (Southeast University), Ministry of Education, China\\
{\tt\small \{fufeng, rhzhou, xieyc, xuyang\_palm, wangjing91, xgeng\}@seu.edu.cn}\\
}

\begin{document}
\twocolumn[{
\maketitle
\begin{figure}[H]

\hsize=\textwidth 
\centering
\vspace{-0.4in}
\includegraphics[width=\textwidth]{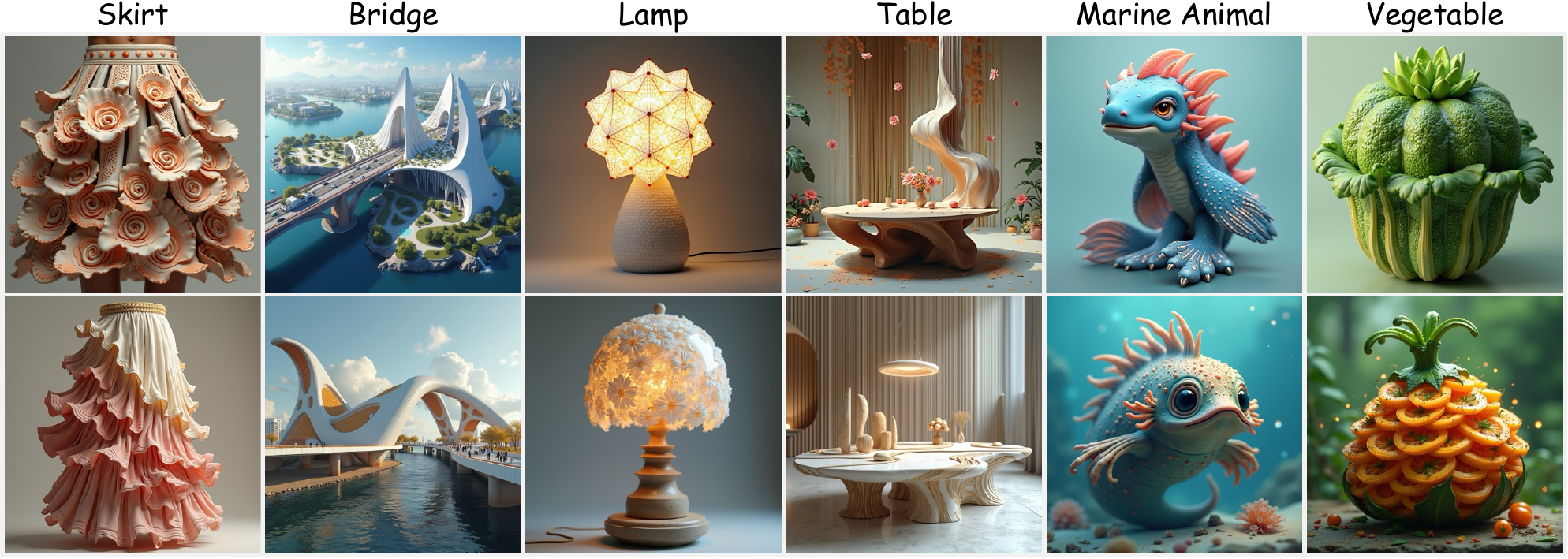}
\vspace{-0.25in}
\caption{\textbf{Breaking Visual Conventions with CosPert.}
CosPert enhances visual creativity through structured perturbations of a pretrained T2I model's weights. Given a category-level prompt such as ``a creative \texttt{[CLS]}'', the perturbed model generates unconventional forms while preserving category identity.}
\label{fig:teaser}
\end{figure}
}]
\renewcommand{\thefootnote}{\fnsymbol{footnote}}
\footnotetext[1]{Equal contribution.}
\footnotetext[2]{Corresponding authors.}
\renewcommand{\thefootnote}{\arabic{footnote}}
\begin{abstract}
Text-to-image (T2I) models often produce high-quality images that follow familiar visual conventions, even in response to open-ended prompts such as ``a creative \texttt{[CLS]}''.
Existing methods encourage novelty by exploring semantic conditions or modifying features and guidance during denoising. 
Pretrained weights offer a complementary source of creative variation, but unstructured perturbations can have little visual effect or degrade image quality. 
We introduce Coupled Spectral Perturbation (CosPert), a training-free method that targets attention projections in layers identified by a one-time sensitivity analysis. 
For each projection, CosPert uses a shared random transformation and its transpose to couple updates to the singular-vector factors while perturbing the spectral gains. 
A common relative budget calibrates the magnitude of weight changes across projections, and a sigmoid schedule concentrates their influence in early denoising and gradually attenuates them during refinement. 
CosPert provides \textbf{plug-and-play} creativity enhancement at inference time. Once the perturbations are prepared, sampling requires no auxiliary networks or additional model evaluations. 
On both FLUX.1 and Stable Diffusion 3.5, CosPert enhances visual creativity, producing more varied object structures while preserving category identity. Across 50 object categories, CosPert achieves the highest mean CREward among the evaluated methods while maintaining strong prompt alignment.
\end{abstract}

\vspace{-0.1in}
\section{Introduction}

\begin{figure*}[t]
  \centering
  \includegraphics[width=\linewidth]{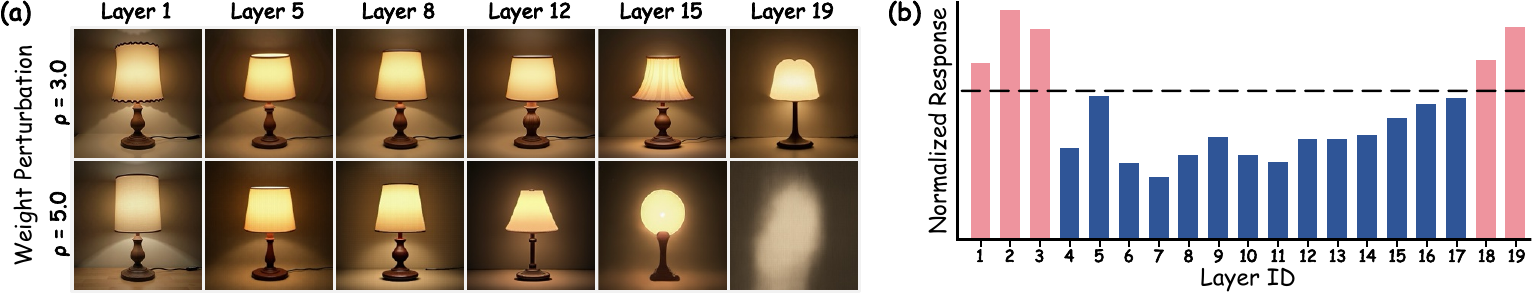}
  \vspace{-0.25in}
  \caption{\textbf{Layer Sensitivity to Weight Perturbation in FLUX.1.}
  (a) We apply Gaussian perturbations one layer at a time at relative magnitudes $\rho=3.0$ and $\rho=5.0$, with $\rho$ defined in Eq.~\eqref{eq:csp_relative_budget}. The images illustrate how the effects vary with the target layer and perturbation strength.
  (b) We rank layers by the normalized change in the backbone prediction and apply CosPert to the five highest-scoring layers.}
  \label{fig:moti}
  \vspace{-0.14in}
\end{figure*}

Text-to-image (T2I) models have made substantial advances in generating high-quality, semantically aligned images from natural language descriptions~\cite{cao2025controllable, shin2025large, huang2025t2i, huberman2026image}. 
These capabilities have expanded their use in artistic exploration~\cite{feng2026breaking, richardson2024conceptlab} and creative design~\cite{zhang2026creatidesign, shi2026creative}, where users seek unexpected yet meaningful visual interpretations that remain coherent and aligned with their intent~\cite{nitzan2026vlm, zhang2025agswap}. 
However, conventional T2I models derive much of their success from approximating complex training data distributions, which can reinforce familiar visual patterns and thereby constrain creative exploration beyond established representations~\cite{dombrowski2025image, adamkiewicz2026pretty, liu2026diversegrpo, basu2026geodiv}.
Consequently, enabling T2I models to break learned visual conventions while maintaining image quality remains a key challenge in creative generation.

Recent approaches enhance generative creativity through semantic exploration~\cite{feng2026breaking, feng2024redefining, richardson2024conceptlab, litp2o} or denoising-time intervention~\cite{han2025enhancing, yu2026self, nitzan2026vlm, lu2024procreate}. 
Semantic exploration targets \textit{combinatorial creativity} by recombining familiar concepts within the semantic space of text conditioning.
BASS~\cite{litp2o} and AGSwap~\cite{zhang2025agswap} fuse concept pairs through sampling and swapping, respectively, while CreTok~\cite{feng2024redefining} and DisTok~\cite{feng2026breaking} explore composition through learned tokens.
Denoising-time interventions instead emphasize \textit{self-creativity}, seeking to generate novel visual instances of a given concept through intermediate feature modulation. C3~\cite{han2025enhancing} amplifies low-frequency feature components, while SCDiff~\cite{yu2026self} applies adaptive spatial weighting to intermediate features.

Despite these advances, existing methods primarily explore conditioning representations and denoising trajectories while keeping pretrained backbone weights fixed. 
However, these weights themselves encode the visual regularities acquired during training and play a central role in shaping the model's preference for familiar patterns. 
This motivates us to investigate \textit{\textbf{whether stochastic weight perturbations can alter these learned preferences and promote creative generation beyond conventional visual patterns.}}

However, perturbing pretrained weights while preserving image quality and semantic coherence is nontrivial. 
As illustrated in Fig.~\ref{fig:moti}a, unstructured random perturbations may leave conventional outputs largely unchanged at low strengths but cause artifacts and semantic drift at higher strengths, with their effects also varying across backbone layers.
These observations suggest that \textit{randomness alone is insufficient for meaningful exploration}. 
Suitable layer selection and structured perturbation design are therefore essential to promote creative variation while preserving the pretrained representations that support coherent image synthesis.

To localize weight-space exploration, we perform a layer-wise sensitivity analysis, illustrated on FLUX.1~\cite{black2024} in Fig.~\ref{fig:moti}b. 
We apply Gaussian perturbations to the attention projections of one layer at a time and rank layers by the resulting change in the backbone prediction. Performed once per pretrained backbone, this analysis narrows the search space to the most sensitive layers.

Having identified the most sensitive layers, we introduce Coupled Spectral Perturbation (CosPert), a training-free method that promotes creative generation through structured perturbations of their attention projection matrices.
CosPert decomposes each matrix into singular-vector factors and spectral gains via singular value decomposition (SVD). It perturbs the gains with a diagonal update and applies \textit{a shared random transformation} and its transpose to opposite sides of the updated spectral core, coupling changes to the input and output directions. A common relative budget controls the magnitude of the resulting weight changes across matrices. 
During generation, a sigmoid schedule concentrates perturbations in early denoising~\cite{habibian2024clockwork, yi2024towards, pan2026semantics}, when object form emerges, and gradually attenuates them so that the pretrained weights increasingly guide refinement.

CosPert offers \textbf{plug-and-play} creativity enhancement at inference time. Once the perturbed weights are prepared, it uses the original architecture with no additional denoising steps or forward passes. 
Qualitative comparisons on FLUX.1 and Stable Diffusion 3.5 show that CosPert breaks learned visual conventions while preserving image quality and semantic coherence.
Across 50 object categories, CosPert achieves the highest mean CREward~\cite{han2026creward} among the evaluated self-creative generation methods and receives the highest preference score in a user study. These results show that structured weight perturbations can enhance creative generation within the backbone's original sampling budget.

Our main contributions are as follows:
1)~We introduce stochastic exploration of pretrained weights for creative image generation, enabling models to break learned visual conventions in response to open-ended prompts.
2)~We propose CosPert, a plug-and-play method for structured exploration of pretrained weights. Layer-wise sensitivity analysis identifies the most responsive layers, where CosPert applies coupled perturbations to the singular-vector factors without additional training.
3)~Extensive experiments demonstrate CosPert's state-of-the-art performance in creative generation, outperforming standard T2I models and creativity-focused methods in creativity and diversity with negligible additional runtime and resource overhead.

\section{Related Work}
Text-to-image (T2I) models have demonstrated remarkable capabilities in synthesizing high-quality, semantically coherent images~\cite{qin2025lumina, zhang2025diffusion, miao2025noise, yu2026pixeldit}. However, generating unexpected yet meaningful visual content remains challenging, particularly in the absence of explicit human guidance~\cite{um2025minority, mateja2021towards, mazzone2019art}.

\vspace{-0.1in}
\paragraph{Combinatorial Creative Generation.}
Combinatorial creativity integrates familiar attributes into novel objects while preserving semantic balance and visual coherence. BASS~\cite{litp2o} and AGSwap~\cite{zhang2025agswap} pair embedding swapping with balance-aware selection and adaptive updates, respectively. RMLer~\cite{li2026rmler} learns mixing policies with compositional balance rewards. For reusable composition, CreTok~\cite{feng2024redefining} learns composition tokens, and CAT~\cite{shi2026creative} derives templates from creative agents. ConceptLab~\cite{richardson2024conceptlab} guides concept discovery with diffusion-prior constraints, while DisTok~\cite{feng2026breaking} uses class-distribution conditioning for concept exploration and composition. CosPert complements these approaches by generating novel designs from a single category prompt through structured stochastic perturbations of pretrained weights.


\vspace{-0.1in}
\paragraph{Self-Creativity in Image Generation.}
Self-creativity aims to generate unexpected yet meaningful images without additional external guidance~\cite{song2026creative}.
Existing approaches promote novelty through sampling guidance or intermediate feature modulation. Guidance-based methods discourage familiar outputs by repelling generated embeddings from references~\cite{lu2024procreate} or adapting negative prompts using VLM feedback on intermediate outputs~\cite{nitzan2026vlm}. 
Feature-based methods instead modulate intermediate representations through low-frequency amplification~\cite{han2025enhancing} or adaptive spatial weighting that balances novelty and semantic alignment~\cite{yu2026self}. Collectively, these approaches promote creative variation by modifying sampling trajectories and intermediate representations.
CosPert instead introduces structured variation directly into pretrained weights to promote creative generation.

\section{Method}
\label{sec:csp_method}

\subsection{Preliminaries}
\label{sec:csp_preliminaries}

\paragraph{Text-Conditioned Generation.}
Given a text condition $c$, a pretrained T2I model iteratively transforms an initial latent $z^{(0)}\sim\mathcal{N}(0,I)$ over $T$ denoising steps. At each step $s$, the sampler updates the latent using the backbone prediction $f_{\theta}$:
\begin{equation}
z^{(s)} =
\mathcal{S}_s\!\left(z^{(s-1)}, f_{\theta}(z^{(s-1)},\tau_s,c)\right),
\quad s=1,\ldots,T,
\label{eq:csp_generation}
\end{equation}
where $\mathcal{S}_s$ is the sampling rule and $\tau_s$ is the scheduled model time. Standard sampling keeps the pretrained parameters $\theta$ fixed and decodes the final latent $z^{(T)}$ into an image.

\vspace{-0.1in}
\paragraph{Problem Formulation.}
Given a pretrained T2I backbone $f_{\theta}$ and a text condition $c$, we aim to generate unexpected yet meaningful images without additional learning or optimization~\cite{feng2026breaking,richardson2024conceptlab}, reference images~\cite{xiong2026vmdiff}, or explicit concept combinations~\cite{litp2o,zhang2025agswap,feng2024redefining}.
We investigate stochastic weight perturbations $\tilde\theta=\theta+\Delta\theta$, sampled independently of the initial latent, to introduce variation directly into the learned transformations while preserving the text condition, initial-noise distribution, and sampler. This extends creative exploration to pretrained weight space, complementing interventions in conditioning representations~\cite{richardson2024conceptlab} and intermediate features~\cite{han2025enhancing,nitzan2026vlm}. The central challenge is \textbf{\textit{how to perturb pretrained weights to promote creative variation while preserving image quality and semantic coherence}}.

\vspace{-0.1in}
\paragraph{Overview.}
To address this challenge, we introduce Coupled Spectral Perturbation (CosPert), a training-free method for structured weight-space exploration. 
We first identify effective perturbation locations through a \textit{one-time} layer-wise sensitivity analysis, narrowing the exploration space (Sec.~\ref{sec:csp_selection}).
Within the selected layers, CosPert applies coupled perturbations to singular-vector factors and spectral gains under a relative weight-change budget (Sec.~\ref{sec:csp_operator}).
A decay schedule concentrates exploration in early denoising and gradually returns the weights toward their pretrained values for refinement (Sec.~\ref{sec:csp_schedule}).

\begin{figure*}[t]
  \centering
  \includegraphics[width=\linewidth]{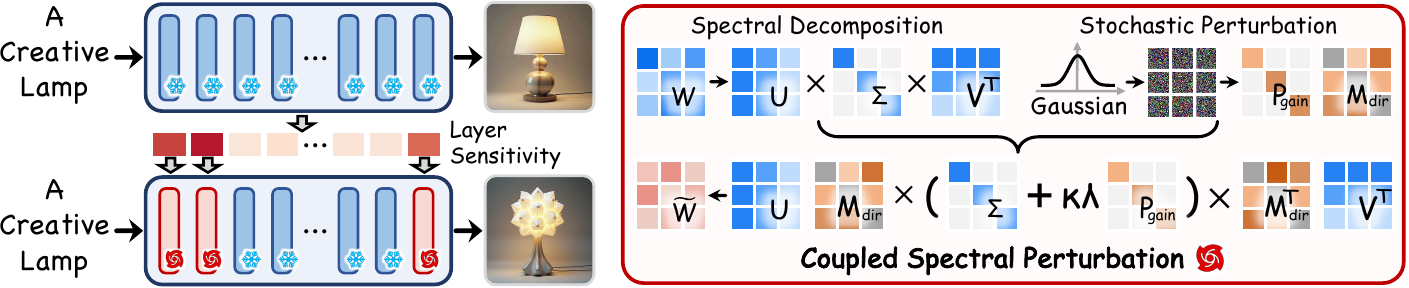}
  \vspace{-0.25in}
  \caption{\textbf{Overview of CosPert.}
  For each selected attention projection, CosPert computes the SVD $W=U\Sigma V^{\top}$ and derives a diagonal gain perturbation $P_{\mathrm{gain}}$ and a directional transformation $M_{\mathrm{dir}}$ from the same Gaussian random matrix. To reconstruct the perturbed weight $\widetilde{W}$, CosPert applies $M_{\mathrm{dir}}$ and $M_{\mathrm{dir}}^{\top}$ to opposite sides of $\Sigma+\kappa\lambda P_{\mathrm{gain}}$, coupling the updates to the input and output directions.}
  \label{fig:method}
  \vspace{-0.1in}
\end{figure*}

\subsection{Layer-wise Sensitivity Analysis}
\label{sec:csp_selection}

The effect of weight perturbation varies across backbone layers, motivating a \textit{one-time} sensitivity analysis to identify effective intervention locations. We focus on attention projections for their central role in governing token interaction and feature aggregation.
For each candidate layer $\ell$, we target its query, key, value, and output projection matrices, denoted by $\mathcal{W}_{\ell}=\{W_{\ell,\star}\mid\star\in\{\mathrm{q},\mathrm{k},\mathrm{v},\mathrm{o}\}\}$.

We probe layers under an empty text condition $c_{\emptyset}$ to reduce prompt dependence. For each $W\in\mathcal{W}_{\ell}$, we independently sample a random scaling factor $\gamma$ and standard Gaussian noise $G$, defining
\begin{equation}
    [\Delta\theta_{\ell}]_W
    =
    (\gamma-1)W
    +\rho_G\|W\|_F\frac{G}{\|G\|_F},
    \label{eq:sensitivity_perturbation}
\end{equation}
where $\rho_G$ controls the relative noise strength.

We perturb one layer at a time and compare original and perturbed predictions at identical latent states and conditions. Let $z^{(s-1)}$ and $y^{(s)}$ denote the unperturbed state and prediction at step $s$. The perturbed prediction is
\begin{equation}
    \tilde y_{\ell}^{(s)}
    =
    f_{\tilde\theta_{\ell}}
    \left(z^{(s-1)},\tau_s,c_{\emptyset}\right),
    \label{eq:csp_probe_prediction}
\end{equation}
where $\tilde\theta_{\ell}=\theta+\Delta\theta_{\ell}$. We quantify each layer's end-to-end influence through relative prediction changes:
\begin{equation}
    S_{\ell}
    =
    \mathbb{E}
    \left[
        \frac{
            \|\tilde y_{\ell}^{(s)}-y^{(s)}\|_2
        }{
            \rho_G\|y^{(s)}\|_2
        }
    \right],
    \label{eq:csp_layer_score}
\end{equation}
where the expectation averages over all denoising steps and repeated trials with independently sampled initial noise and weight perturbations. The highest-scoring layers form the intervention set $\mathcal{L}$, selected \textbf{\textit{once per pretrained backbone}} and reused across generations.

\subsection{Coupled Spectral Perturbation}
\label{sec:csp_operator}

With the intervention layers $\mathcal{L}$ identified, we use SVD to separate each selected matrix $W\in\mathcal{W}_{\ell}$ into directional factors and spectral gains:
\begin{equation}
    W=U\Sigma V^{\top},
    \qquad
    \Sigma=\operatorname{Diag}(\sigma_1,\ldots,\sigma_r),
    \label{eq:csp_svd}
\end{equation}
where $W\in\mathbb{R}^{d_{\mathrm{out}}\times d_{\mathrm{in}}}$ and $r=\min(d_{\mathrm{out}},d_{\mathrm{in}})$. The factors $V$ and $U$ specify input and output directions, respectively, while $\Sigma$ controls their gains. This decomposition enables coordinated perturbations of directions and gains.

To coordinate perturbations of directional factors and spectral gains, we sample a standard Gaussian matrix $G\in\mathbb{R}^{r\times r}$ and construct two normalized perturbation directions:
\begin{equation}
    P_{\mathrm{dir}}
    =
    \frac{G-G^{\top}}{\|G-G^{\top}\|_F},
    \;\;
    P_{\mathrm{gain}}
    =
    \operatorname{Diag}\!\left(
        \frac{\operatorname{diag}(G)}
        {\|\operatorname{diag}(G)\|_2}
    \right).
    \label{eq:csp_random_directions}
\end{equation}
Here, $P_{\mathrm{dir}}$ governs singular-vector mixing, while $P_{\mathrm{gain}}$ controls spectral gain perturbation.

As illustrated in Fig.~\ref{fig:method}, we reconstruct the perturbed weight by applying the shared directional transformation $M_{\mathrm{dir}}=I_r+\lambda P_{\mathrm{dir}}$ and its transpose to opposite sides of the updated spectral core, yielding
\begin{equation}
    \widetilde W
    =
    UM_{\mathrm{dir}}
    \left(\Sigma+\kappa\lambda P_{\mathrm{gain}}\right)
    M_{\mathrm{dir}}^{\top}V^{\top},
    \label{eq:csp_congruence}
\end{equation}
where $\lambda\geq0$ controls the overall perturbation strength and $\kappa$ scales the gain perturbation relative to the directional update.
This construction couples the updates to the two singular-vector factors, and the full matrix product retains their interactions with the gain perturbation.

To ensure comparable relative weight changes across matrices, we calibrate $\lambda$ independently for each $W$. The resulting update $\Delta W=\widetilde W-W$ must satisfy
\begin{equation}
    \frac{\|\Delta W\|_F}{\|W\|_F}=\rho,
    \label{eq:csp_relative_budget}
\end{equation}
where $\rho$ specifies the target relative strength. We solve this constraint through a one-dimensional numerical search over $\lambda$, keeping $P_{\mathrm{dir}}$ and $P_{\mathrm{gain}}$ fixed.

\begin{figure*}[t]
  \centering
  \includegraphics[width=\linewidth]{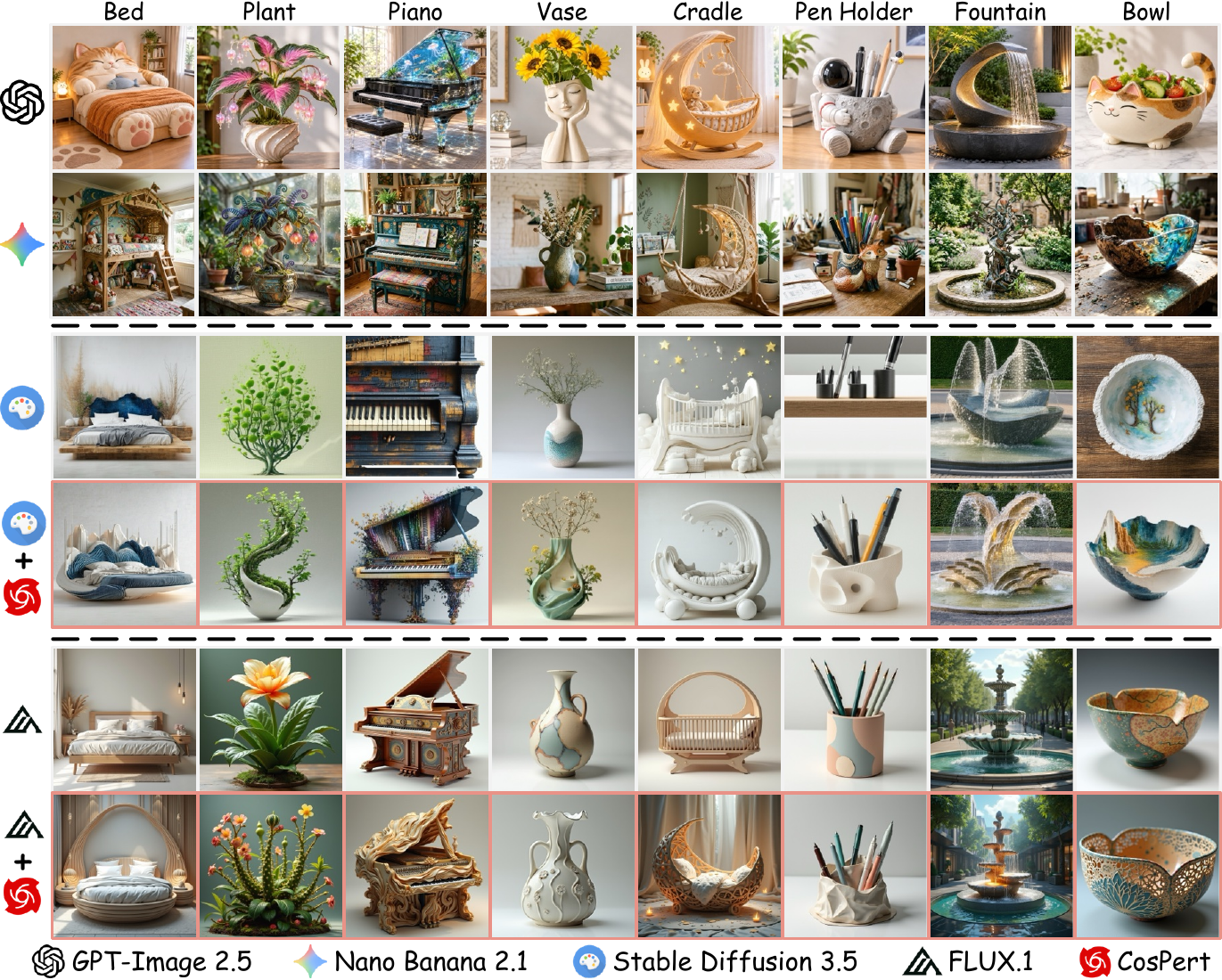}
  \vspace{-0.22in}
  \caption{\textbf{Qualitative Comparison with State-of-the-Art T2I Models.}
  We compare GPT-Image 2.5, Nano Banana 2.1, Stable Diffusion 3.5, and FLUX.1. For Stable Diffusion 3.5 and FLUX.1, we show paired outputs generated with the original weights and with CosPert.}
  \label{fig:main_t2i}
  \vspace{-0.1in}
\end{figure*}

\begin{figure*}[t]
  \centering
  \includegraphics[width=\linewidth]{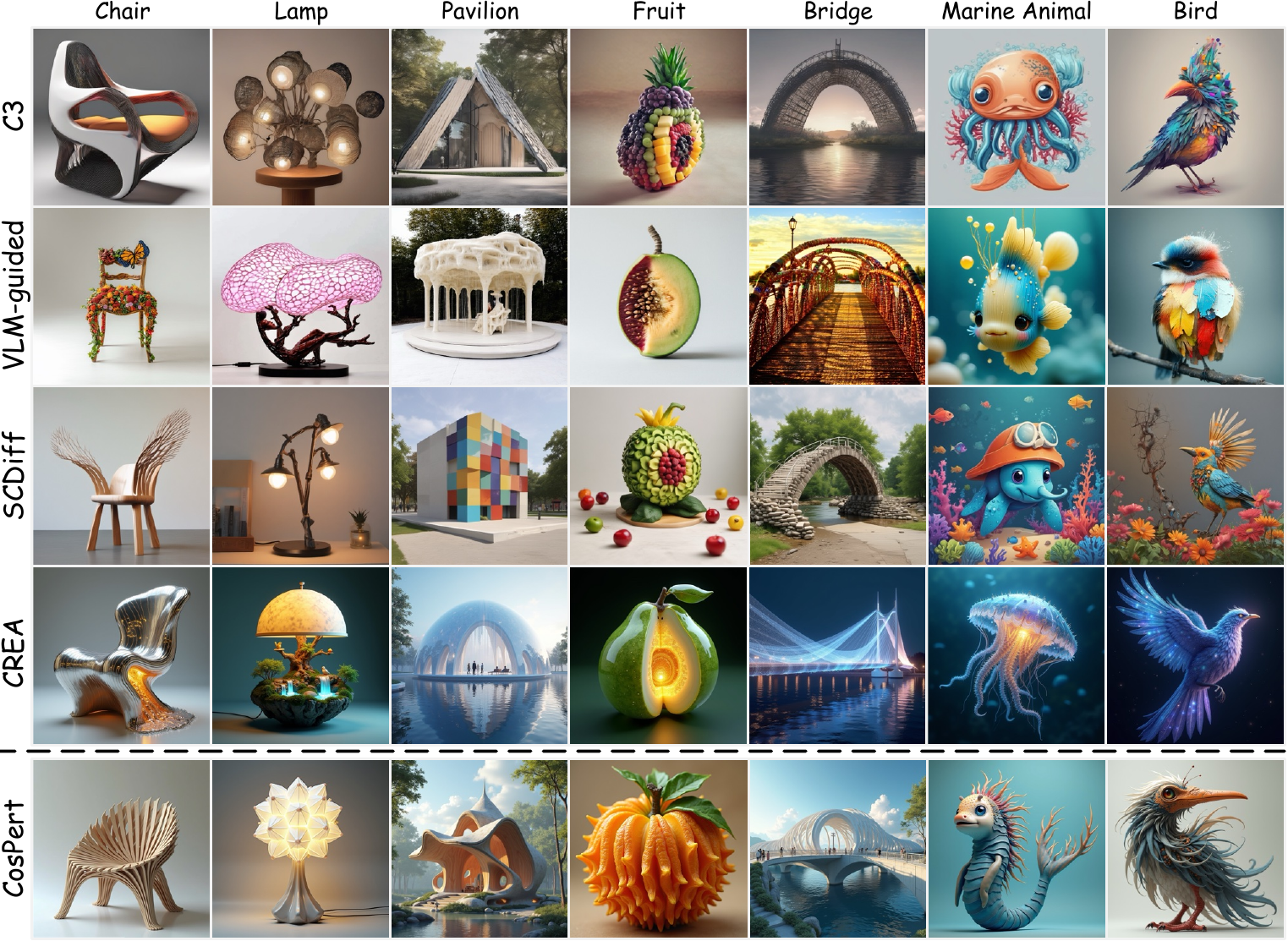}
  \vspace{-0.25in}
  \caption{\textbf{Qualitative Comparison with Self-Creative Generation Methods.}
  We compare CosPert with C3~\cite{han2025enhancing} and SCDiff~\cite{yu2026self}, which modulate intermediate features during denoising, as well as VLM-guided~\cite{nitzan2026vlm} and CREA~\cite{venkateshcrea}, which derive creative guidance from a vision--language model and an LLM-based agent, respectively.}
  \label{fig:main_crea}
  \vspace{-0.05in}
\end{figure*}

\subsection{Temporally Scheduled Exploration}
\label{sec:csp_schedule}

With perturbation magnitudes calibrated, we schedule their influence during denoising to encourage early exploration and support later refinement. Let $\mathcal{W}_{\mathcal{L}}=\bigcup_{\ell\in\mathcal{L}}\mathcal{W}_{\ell}$ denote the selected projections. Their calibrated updates define a fixed full-model perturbation:
\begin{equation}
    [\Delta\theta]_W
    =
    \begin{cases}
        \Delta W, & W\in\mathcal{W}_{\mathcal{L}},\\
        0, & \text{otherwise}.
    \end{cases}
    \label{eq:csp_full_perturbation}
\end{equation}
At denoising step $s$, we scale the fixed perturbation by a decay coefficient $d_s$:
\begin{equation}
    \tilde\theta_s
    =
    \theta+d_s\Delta\theta.
    \label{eq:csp_scheduled_parameters}
\end{equation}
To transition smoothly from early exploration to later refinement, we use a sigmoid decay schedule:
\begin{equation}
d_s=
\frac{1}{1+\exp\!\left(\alpha(s-m)\right)},
\qquad s=1,\ldots,T,
\label{eq:csp_schedule}
\end{equation}
where $m$ sets the decay midpoint and $\alpha>0$ controls its steepness. As $d_s$ decreases during denoising, the update $\widetilde W_s=W+d_s\Delta W$ progressively reduces the relative perturbation strength to $d_s\rho$ while preserving its direction.

Overall, CosPert enables training-free creative generation through structured, temporally scheduled weight perturbations, requiring no auxiliary networks or additional model evaluations during sampling.

\section{Experimental Setup}
\label{sec:exp}

\paragraph{Dataset.}
To evaluate self-creativity, we construct a benchmark of 50 common categories spanning manufactured and natural objects. For each category, we use a short text prompt as the sole conditioning input, such as ``A photorealistic \texttt{[CLS]} with a creative visual design.'' The prompt specifies the target category while leaving the design choices to the model. Because CosPert is \textbf{\textit{training-free}}, all 50 categories are used directly for evaluation.

\vspace{-0.15in}
\paragraph{Implementation Details.}
We use FLUX.1~\cite{black2024} as the primary backbone with $T=28$ denoising steps. We estimate layer sensitivity once using 64 sampling trajectories with $\gamma\sim\mathcal{U}(0.90,1.10)$ and $\rho_G=1.5$, then select the five highest-scoring layers for perturbation. For generation, we set $\rho=1.0$ and $\kappa=0.7/1.5$ and use a sigmoid decay schedule with $m=6$ and $\alpha=4$. CosPert is training-free and constructs perturbations independently for each concept, with no additional model evaluations during sampling. 
\section{Results}
\subsection{Qualitative Evaluation}
\paragraph{Comparison with State-of-the-Art T2I Models.}

Fig.~\ref{fig:main_t2i} compares outputs for ``A photorealistic \texttt{[CLS]} with a creative visual design.'', which specifies the object category and leaves the design open.
GPT-Image 2.5 and Nano Banana 2.1 appear to infer what could make each object creative, often expressing distinct themes through motifs, surface treatments, or scene composition while retaining familiar structures.
Stable Diffusion 3.5 and FLUX.1 show a weaker response to the same cue, typically producing conventional forms with modest decorative variation.

Given the same prompt, CosPert enhances visual creativity in both Stable Diffusion 3.5 and FLUX.1. Each perturbation is constructed in the attention projection's own singular-vector basis, with its magnitude calibrated relative to the original weight norm. 
This construction accounts for differences in weight spectra and scales, allowing the same plug-and-play procedure to work across both backbones.

\vspace{-0.14in}
\paragraph{Comparison with Self-Creative Methods.}
Fig.~\ref{fig:main_crea} compares self-creative generation methods across familiar categories. 
C3~\cite{han2025enhancing} and SCDiff~\cite{yu2026self} modulate intermediate features during denoising, often varying surface details or individual parts. 
Pronounced changes can reduce visual realism, as illustrated by the exaggerated proportions of SCDiff's marine animal. 
The VLM-guided method~\cite{nitzan2026vlm} uses visual feedback to discourage conventional elements, while CREA~\cite{venkateshcrea} uses an LLM agent to infer a design direction from the prompt. In the examples shown, both methods express novelty primarily through ornamentation, materials, or lighting, with limited changes to overall geometry.

CosPert couples singular-vector updates and adjusts spectral gains in pretrained weights. These perturbations alter feature aggregation during early denoising, promoting novel silhouettes and part arrangements from a category prompt. As they decay, the pretrained weights increasingly guide refinement and help preserve category identity.

\begin{table*}[t]
    \centering
    \setlength{\tabcolsep}{1.6mm}
    \caption{\textbf{Quantitative Comparison of Creative Generation.}
    \textbf{(a)} CREward~\cite{han2026creward} is a reward model trained specifically to assess image creativity.
    \textbf{(b)} VQAScore~\cite{lin2024evaluating} measures text--image alignment, and PickScore~\cite{kirstain2023pick} estimates human preference.
    \textbf{(c)} User study results are reported as Plackett--Luce scores~\cite{plackett1975analysis} estimated from participants' rankings. 
    Values following $\pm$ denote standard deviations.}
    \vspace{-0.1in}
    \resizebox{\textwidth}{!}{%
    \begin{tabular}{@{}llcccc|cccc|c@{}}
        \toprule
        & & FLUX.1 & SD3.5 & GPT & Nano Banana
          & C3 & VLM-Guide & SCDiff & CREA & CosPert \\
        \midrule
        \textbf{(a)} & CREward$\uparrow$
        & $-0.70_{\pm 1.63}$
        & $1.12_{\pm 2.60}$
        & $0.87_{\pm 2.18}$
        & $3.02_{\pm 2.10}$
        & $2.33_{\pm 2.32}$
        & $2.00_{\pm 2.55}$
        & $1.40_{\pm 2.07}$
        & $3.15_{\pm 1.86}$
        & $\mathbf{3.69}_{\pm 1.50}$ \\
        \midrule
        \multirow{2}{*}{\textbf{(b)}} & VQAScore$\uparrow$
        & $0.83_{\pm 0.06}$
        & $0.83_{\pm 0.07}$
        & $0.88_{\pm 0.06}$
        & $0.84_{\pm 0.09}$
        & $0.83_{\pm 0.10}$
        & $0.84_{\pm 0.06}$
        & $0.82_{\pm 0.07}$
        & $0.87_{\pm 0.05}$
        & $\mathbf{0.89}_{\pm 0.04}$ \\
        & PickScore$\uparrow$
        & $21.81_{\pm 0.68}$
        & $21.80_{\pm 0.94}$
        & $22.26_{\pm 0.79}$
        & $22.14_{\pm 0.83}$
        & $21.74_{\pm 0.76}$
        & $21.46_{\pm 0.80}$
        & $21.53_{\pm 0.72}$
        & $21.98_{\pm 1.00}$
        & $\mathbf{22.26}_{\pm 0.73}$ \\
        \midrule
        \textbf{(c)} & User Study$\uparrow$
        & $2.84_{\pm 0.70}$
        & $5.83_{\pm 1.23}$
        & $16.69_{\pm 2.96}$
        & $10.10_{\pm 1.94}$
        & $5.87_{\pm 1.28}$
        & $6.44_{\pm 1.39}$
        & $5.19_{\pm 1.07}$
        & $15.01_{\pm 2.51}$
        & $\mathbf{32.03}_{\pm 4.54}$ \\
        \bottomrule
    \end{tabular}%
    }
\vspace{-0.1in}
\label{tab:eval}
\end{table*}

\begin{figure*}[t]
    \centering
    \begin{minipage}[c]{0.44\linewidth}
        \centering
        \setlength{\tabcolsep}{1.5pt}
        \resizebox{\linewidth}{!}{%
        \begin{tabular}{@{}lccc|ccc@{}}
            \toprule
            & $P_{\mathrm{gain}}$ & $P_{\mathrm{dir}}$ & Coupled
            & CREward & VQASc. & PickSc. \\
            \midrule
            FLUX.1 (w/o Pert.)   & --         & --         & --         & $-0.70$ & $0.83$ & $21.81$ \\
            Gaussian Pert.      & --         & --         & --         & $0.02$  & $0.82$ & $21.83$ \\
            \midrule
            w/o Direction Pert. & \checkmark & --         & --         & $0.46$  & $0.82$ & $21.99$ \\
            w/o Gain Pert.      & --         & \checkmark & \checkmark & $3.04$  & $0.87$ & $22.03$ \\
            w/o Coupling        & \checkmark & \checkmark & --         & $0.81$  & $0.84$ & $21.96$ \\
            \midrule
            \rowcolor{gray!15}
            \textbf{CosPert}    & \checkmark & \checkmark & \checkmark & $\mathbf{3.69}$ & $\mathbf{0.89}$ & $\mathbf{22.26}$ \\
            \bottomrule
        \end{tabular}%
        }
    \end{minipage}\hfill
    \begin{minipage}[c]{0.54\linewidth}
        \centering
        \includegraphics[width=\linewidth]{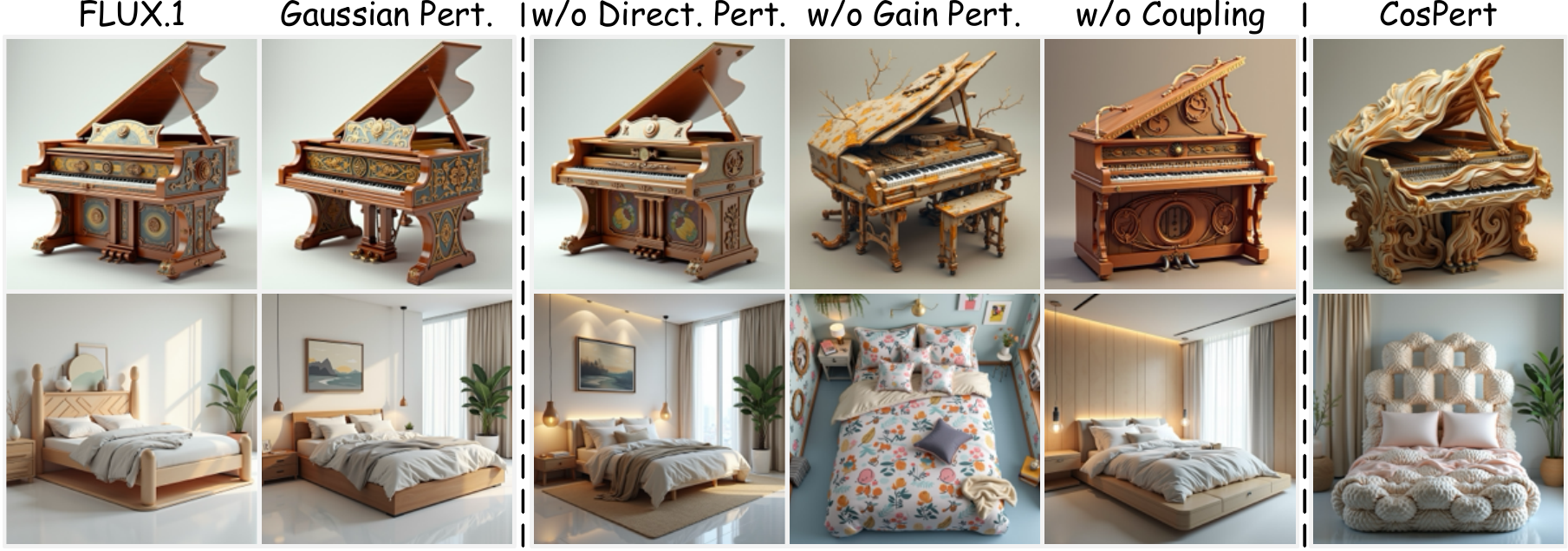}
    \end{minipage}
    \vspace{-0.1in}
    \caption{\textbf{Ablation of Coupled Spectral Perturbation.}
    The table compares unperturbed FLUX.1, additive Gaussian perturbations, and CosPert variants on CREward, VQAScore, and PickScore. Representative outputs for each setting appear on the right.}
    \vspace{-0.05in}
    \label{fig:ablation_spectral}
\end{figure*}

\begin{figure*}[!t]
    \centering
    \begin{minipage}[c]{0.27\linewidth}
        \centering
        \setlength{\tabcolsep}{2pt}
        \resizebox{\linewidth}{!}{%
            \begin{tabular}{@{}lccc@{}}
                \toprule
                & CREward & VQASc. & PickSc. \\
                \midrule
                Bottom-5  & $0.03$ & $0.82$ & $21.91$ \\
                Uniform-5 & $0.95$ & $0.84$ & $22.10$ \\
                Random-5  & $0.12$ & $0.81$ & $21.87$ \\ 
                \midrule
                Top-1 & $1.01$ & $0.84$ & $22.00$ \\
                Top-3 & $1.47$ & $0.84$ & $22.08$ \\
                Top-7 & $2.40$ & $0.86$ & $22.00$ \\
                \midrule
                \rowcolor{gray!15}
                \textbf{Top-5 (Ours)} & $\mathbf{3.69}$ & $\mathbf{0.89}$ & $\mathbf{22.26}$ \\
                \bottomrule
            \end{tabular}%
        }
    \end{minipage}\hfill
    \begin{minipage}[c]{0.71\linewidth}
        \centering
        \includegraphics[width=\linewidth]{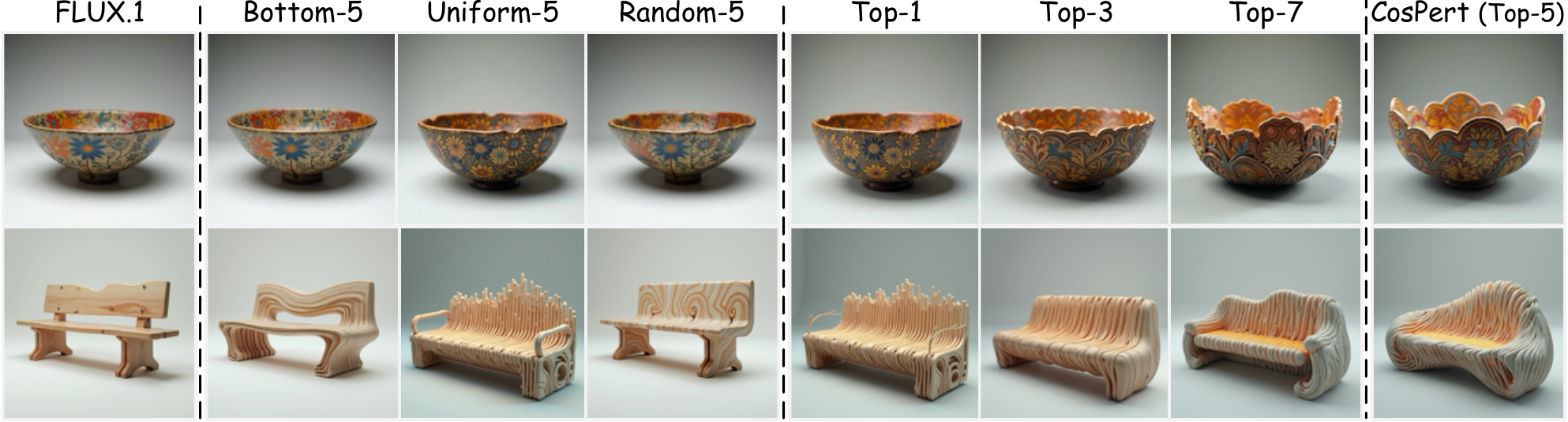}
    \end{minipage}
    \vspace{-0.1in}
    \caption{\textbf{Ablation of Layer Selection.}
    The table on the left compares different numbers of the highest-sensitivity layers and different strategies for selecting five layers, using unperturbed FLUX.1 as the baseline. The right shows representative generations.}
    \vspace{-0.05in}
    \label{fig:ablation}
\end{figure*}

\subsection{Quantitative Evaluation}

We compare CosPert with pretrained T2I models and methods designed for creative generation, evaluating visual creativity, text--image alignment, and human preference.

\vspace{-0.15in}
\paragraph{Creativity Reward Model}
We evaluate image creativity with CREward~\cite{han2026creward}, a reward model trained to assess novelty in geometry, materials, and textures. CosPert achieves the highest mean score in Table~\ref{tab:eval}(a), consistent with qualitative evidence that coupled spectral perturbations yield more varied object forms and appearances.

\vspace{-0.18in}
\paragraph{Automated Vision-language Evaluation}
We measure text--image alignment with VQAScore~\cite{lin2024evaluating} and estimate human preference with PickScore~\cite{kirstain2023pick}. 
CosPert achieves the highest mean scores on both metrics (Table~\ref{tab:eval}(b)), with its VQAScore indicating that structured weight perturbations preserve semantic alignment while enhancing creativity.

\vspace{-0.18in}
\paragraph{User Study.}
To assess human preference directly, we ask participants to rank images from the compared methods and fit a Plackett--Luce model~\cite{plackett1975analysis} to their rankings. CosPert achieves the highest estimated preference score in Table~\ref{tab:eval}(c), complementing the automated evaluation with direct human judgments.

\begin{figure*}[!t]
  \centering
  \includegraphics[width=\linewidth]{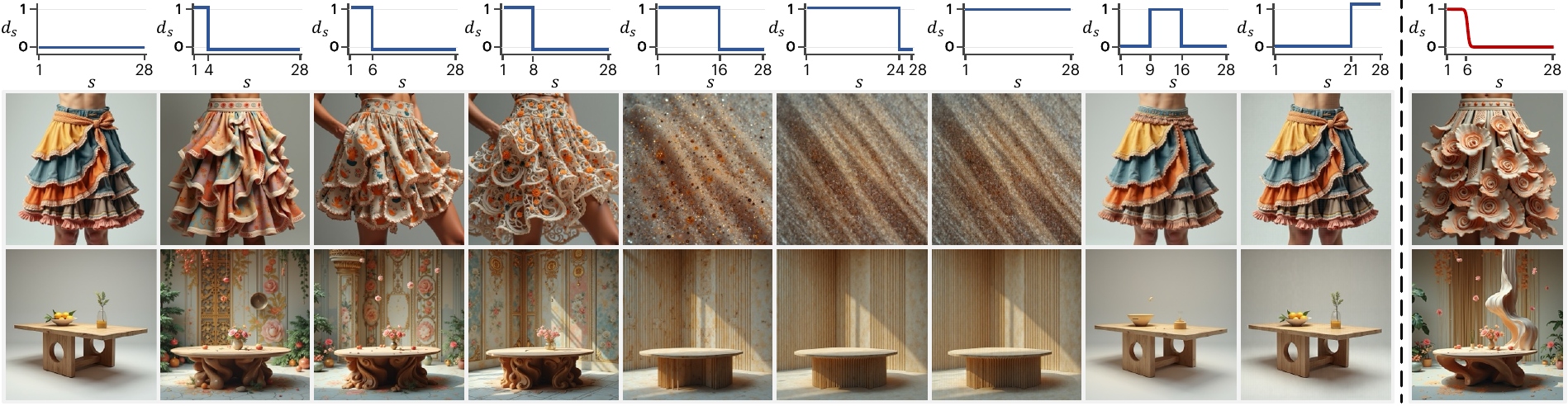}
  \vspace{-0.25in}
  \caption{\textbf{Effect of Perturbation Timing on Creative Generation.}
  The curves plot the perturbation weight $d_s$ at denoising step $s$. The images compare perturbation windows of varying onset and duration with CosPert's sigmoid decay schedule.}
  \label{fig:ana_decay}
  \vspace{-0.14in}
\end{figure*}

\begin{figure*}[t]
  \centering
  \includegraphics[width=\linewidth]{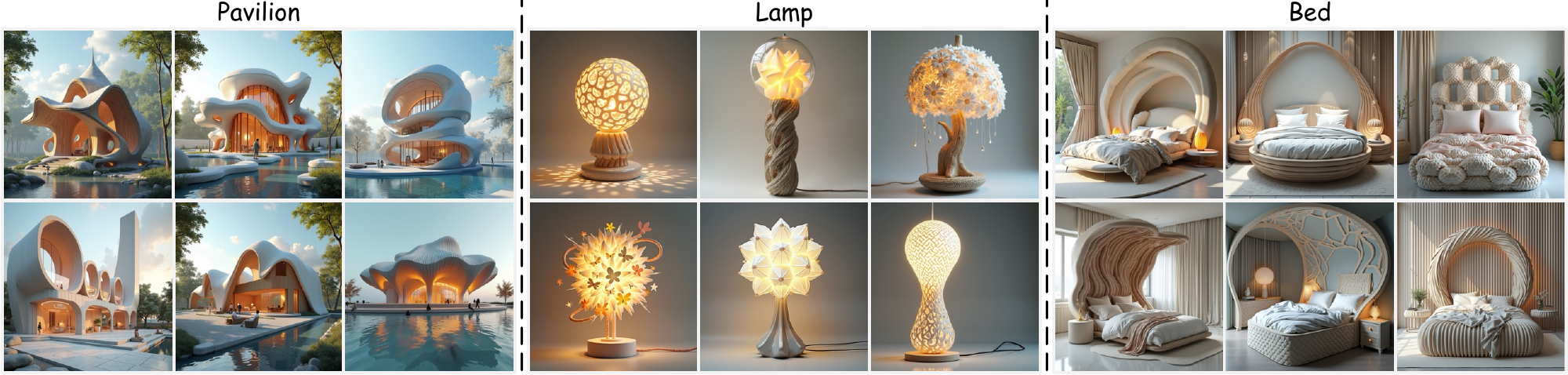}
  \vspace{-0.27in}
   \caption{\textbf{Creative Diversity across Perturbation Seeds.}
    For each category, CosPert generates six distinct designs by varying the weight perturbation seed under a fixed text prompt.}
  \label{fig:ana_div}
  \vspace{-0.1in}
\end{figure*}

\begin{figure*}[!t]
  \centering
  \includegraphics[width=\linewidth]{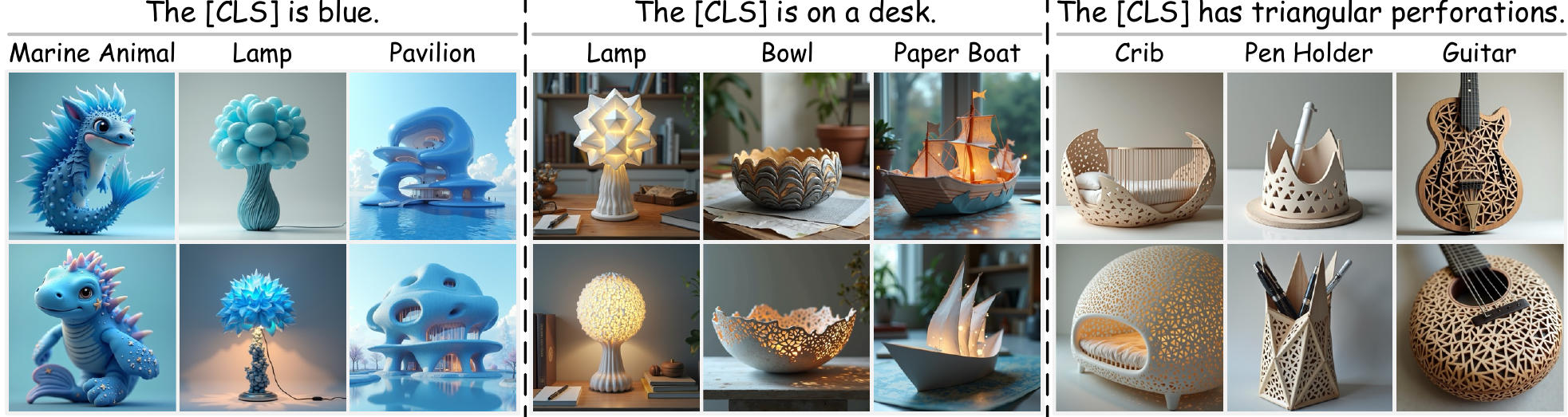}
  \vspace{-0.29in}
  \caption{\textbf{Preserving Prompt Constraints in Creative Generation.}
  CosPert's weight perturbations introduce structural variation while preserving prompt-specified constraints such as color, object placement, and perforation shape.}
  \label{fig:ana_sema}
  \vspace{-0.14in}
\end{figure*}

\subsection{Ablation and Analysis}
\subsubsection{Ablation on Coupled Spectral Perturbation}
Fig.~\ref{fig:ablation_spectral} isolates the effects of gain perturbation, directional perturbation, and coupling.
Gain perturbation alone (\textit{w/o Direction Pert.}) yields only a modest CREward improvement over unperturbed FLUX.1, while coupled directional updates alone (\textit{w/o Gain Pert.}) retain much of CosPert's improvement.
Removing coupling (\textit{w/o Coupling}) also lowers CREward even when both perturbation components are present.
CosPert uses a shared transformation to couple changes in the input and output directions, while gain perturbations adjust the strength of individual spectral modes. Together, these updates may explain how CosPert generates more varied object shapes while preserving category identity.

\subsubsection{Ablation on Sensitivity-Based Layer Selection}
Fig.~\ref{fig:ablation} examines how the choice and number of perturbed layers affect generation. 
At a fixed budget of five layers, Top-5 outperforms alternative selection strategies across all metrics (e.g., Uniform-5). Scores peak at Top-5 and decline at Top-7, possibly because the effects of perturbing additional, less sensitive layers accumulate across the backbone and disrupt features that support prompt alignment and visual coherence. In the qualitative examples, both settings yield pronounced geometric changes while preserving category identity.

\subsubsection{Ablation of the Temporal Perturbation Schedule}
Fig.~\ref{fig:ana_decay} examines how the onset and duration of weight perturbations affect generation.
Early perturbations reshape the skirt silhouette and table base, while late interventions produce results close to the unperturbed baseline. 
This suggests that overall geometry is largely established early in denoising.
Sustaining perturbations into later steps can hinder refinement, as illustrated by the repetitive textures that obscure the skirt's silhouette.
CosPert therefore uses sigmoid decay to emphasize structural variation early and gradually return the weights toward their pretrained values during refinement.

\subsubsection{Analysis of Diversity across Perturbation Seeds}
Fig.~\ref{fig:ana_div} shows that varying the perturbation seed for a fixed category prompt produces distinct pavilion roofs, lamp silhouettes, and bed headboards while preserving category identity.  CosPert calibrates spectral updates to a common relative magnitude, so changing the seed varies the direction of the weight perturbation and alters how attention aggregates features during early denoising. Setting the perturbation seed independently of the initial latent seed provides a separate way to control structural diversity.

\subsubsection{Analysis of Semantic Preservation}
Fig.~\ref{fig:ana_sema} examines whether the structural variation introduced by CosPert remains consistent with explicit prompt constraints. Across paired generations, objects retain their category identity and satisfy the specified attributes and spatial relations as their silhouettes and part arrangements change. The perforation examples are particularly informative because the prompt constrains geometry at the local level. The openings remain triangular across changes in overall form, suggesting that weight perturbations can expand the range of designs along dimensions left open by the prompt.

\section{Conclusion}
We introduced CosPert, a plug-and-play method for breaking learned visual conventions through structured perturbations of pretrained weights.
A one-time sensitivity analysis identifies the layers where CosPert couples singular-vector updates and perturbs the spectral gains of attention projections.
Once the perturbations are prepared, CosPert operates within the backbone's original sampling budget.
Experiments on FLUX.1 and Stable Diffusion 3.5 show that CosPert substantially enhances visual creativity, producing more varied object structures while preserving category identity. 

\section*{LLM Usage Statement}

The authors used generative AI tools, specifically GPT and Codex powered by GPT models, to assist with writing, LaTeX and figure formatting, and code editing. All claims, citations, code, figures, and numerical results were independently reviewed and verified by the authors, who take full responsibility for the final submission.

{
    \small
    \bibliographystyle{ieeenat_fullname}
    \bibliography{main}
}


\end{document}